\documentclass[twocolumn, switch]{article} % Method A for two-column formatting

\usepackage{preprint}

\usepackage{graphics} % for pdf, bitmapped graphics files
\usepackage{epsfig} % for postscript graphics files
\usepackage{mathptmx} % assumes new font selection scheme installed
\usepackage{times} % assumes new font selection scheme installed
\usepackage{amsmath} % assumes amsmath package installed
\usepackage{amssymb}  % assumes amsmath package installed
\usepackage[ruled,vlined]{algorithm2e}
\usepackage{booktabs}
\usepackage{tikz}
\usepackage{array}
\usetikzlibrary{arrows,positioning,fit,shapes,calc}
\usepackage{multirow}
\usepackage{multicol}
\usepackage{subcaption}
\usepackage{mwe}
\usepackage{amssymb}
\usepackage{paralist}
\usepackage{pifont}% http://ctan.org/pkg/pifont
\newcommand{\cmark}{\ding{51}}%
\newcommand{\xmark}{\ding{55}}%
\usepackage{fdsymbol}

\newcommand{\eg}{\emph{e.g.}, }
\newcommand{\ie}{\emph{i.e.}, } 

\title{Semantically-Aware Strategies for Stereo-Visual Robotic Obstacle Avoidance}

\author{Jungseok Hong$^1$, Karin de Langis$^2$, Cole Wyeth$^3$, Christopher Walaszek$^4$, and Junaed Sattar$^5$
\thanks{The authors are with the Department of Computer Science and Engineering, Minnesota Robotics Institute, University of Minnesota--Twin Cities, 100 Union St SE, Minneapolis, MN, 55455, USA. {\tt\small \{$^1$jungseok,$^2$dento019,$^3$wyeth008,
$^4$walas013,$^5$junaed\}@umn.edu.}}
\thanks{*This work was supported by the US National Science Foundation awards IIS-\#1637875 \& IIS-\#1845364, the UMII-MnDRIVE Fellowship, the MnRI Seed Grant, and Nvidia GPU Grant.}
}

\begin{document}

\twocolumn[ % Method A for two-column formatting
  \begin{@twocolumnfalse} % Method A for two-column formatting

\maketitle
\thispagestyle{empty}
\pagestyle{empty}

\begin{abstract}
Mobile robots in unstructured, mapless environments must rely on an obstacle avoidance module to navigate safely. The standard avoidance techniques estimate the locations of obstacles with respect to the robot but are unaware of the obstacles’ identities. Consequently, the robot cannot take advantage of semantic information about obstacles when making decisions about how to navigate. We propose an obstacle avoidance module that combines visual instance segmentation with a depth map to classify and localize objects in the scene. The system avoids obstacles differentially, based on the identity of the objects: for example, the system is more cautious in response to unpredictable objects such as humans. The system can also navigate closer to harmless obstacles and ignore obstacles that pose no collision danger, enabling it to navigate more efficiently. We validate our approach in two simulated environments: one terrestrial and one underwater. Results indicate that our approach is feasible and can enable more efficient navigation strategies.
\end{abstract}

\vspace{0.35cm}

  \end{@twocolumnfalse} % Method A for two-column formatting
] % Method A for two-column formatting

\section{INTRODUCTION}
An autonomous robot navigating an unstructured, mapless environment typically avoids obstacles by utilizing range information received from its sensors. When a danger of collision arises, the robot adjusts its trajectory to ensure safety. The state-of-the-art techniques in obstacle avoidance effectively prevent collisions, but they are unaware of the semantic properties of the obstacles they are avoiding \cite{hoy2015algorithms}. Consequently, they are unable to exploit a more in-depth understanding of the objects in the environment to navigate more efficiently without compromising safety. In this paper, we augment an obstacle avoidance module's capabilities by incorporating semantic information and show experimentally that this results in more efficient navigation. We refer to this augmentation as Semantic Obstacle Avoidance for Robots (SOAR).
% \begin{figure}[hbt!]
%  \centering
% \includegraphics{example-image-a}
% \end{figure}

\begin{figure}
    \vspace{3mm}
    \centering
    % \missingfigure[figwidth=0.49\textwidth]{test}
    \includegraphics[width=\linewidth]{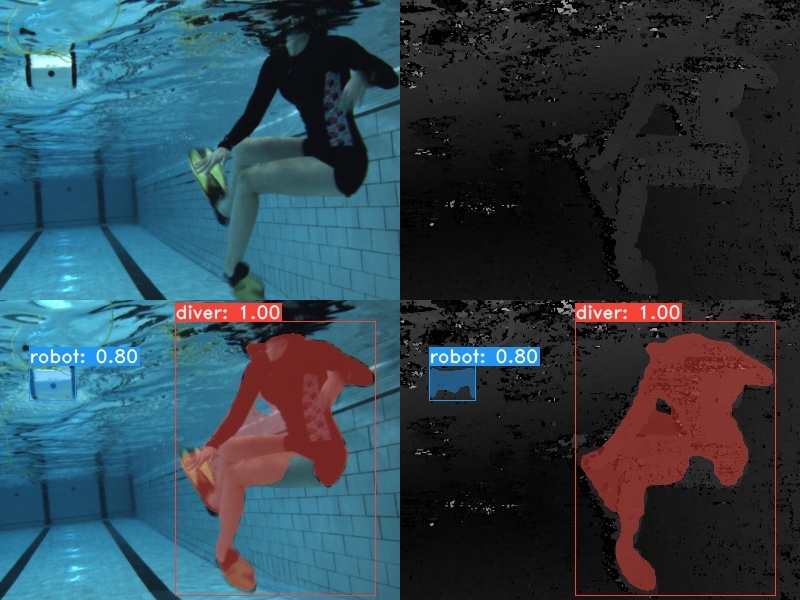}
    \caption{Four levels of scene understanding in underwater human-robot collaborative missions, shown top-to-bottom, left-to-right: observation with no understanding (RGB image), distance-aware observation without scene understanding (depth estimation), instance-aware scene understanding (instance segmentation), and depth-instance-aware scene understanding (depth-instance segmentation). We propose a depth-instance-aware approach for obstacle avoidance.}
    \label{fig:intofig}
  \vspace{-4mm}
\end{figure}

Semantic information is often used to aid robot navigation in structured environments \cite{kostavelis2015semantic, galindo2008robot}. In this work, we propose that semantic information can also be useful for obstacle avoidance in unstructured environments. For example, a robot may want to consider that a living thing, like a person or a dog, has the potential to start moving, even if it is currently stationary. The robot may also want to consider that some obstacles may not actually pose a collision danger: for instance, a depth map may detect plastic balls in the robot's direction of motion, but the robot can safely collide with the balls. When the robot is able to recognize that different objects pose different collision dangers, it can choose a path that maximizes efficiency without jeopardizing safety. In fact, semantic obstacle avoidance is desirable in several robotic applications with unstructured environments, including:
\begin{itemize}
    \item \textbf{Robotic wheelchairs} need to generally stay clear of obstacles, but they may want to give extra clearance to objects like doors that have the potential to suddenly swing forward. On the other hand, if the user wants to dock at a table, the wheelchair needs to allow itself to get very close to it.
    \item \textbf{Autonomous underwater vehicles} (AUVs) are employed by marine biologists to observe endangered species of mussels, which are usually embedded within rock formations, requiring AUVs to get much closer.
    \item \textbf{Diver-following AUVs} need to avoid obstacles while recognizing bubbles emanating from the diver's flippers do not actually pose a collision danger.
\end{itemize}
\begin{figure}
    \vspace{3mm}
    \centering
    % \missingfigure[figwidth=0.49\textwidth]{test}
    \includegraphics[width=\linewidth]{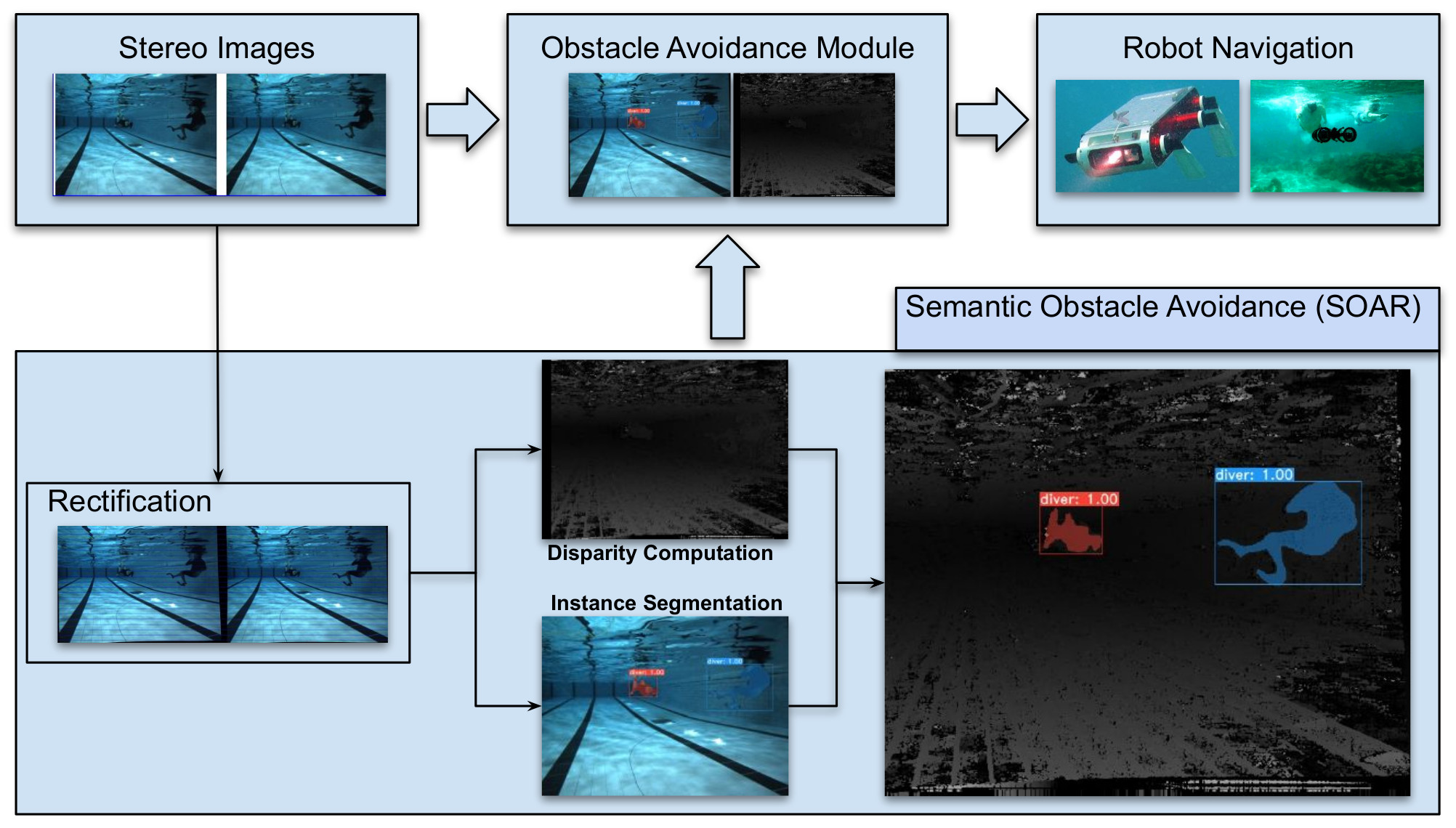}
    \caption{Illustration of our obstacle avoiding approach which fuses depth and semantic information for selective avoidance. The input is a pair of stereo images, which is used to both compute disparity and, through YOLACT, generate semantic labels for obstacles in the scene. Fusing these, a robot has both depth estimates and semantic information about potential obstacles, enabling it to select navigation strategies depending on the nature of the obstacle.}
    \label{fig:pipeline}
    \vspace{-4mm}
\end{figure}

It is imperative to note that even if a map of the environment is available, the presence of dynamic objects (\eg in the above scenarios, people, wheelchairs, and fish) will require information beyond the ``free-and-occluded'' labels that are usually provided by maps. Semantic maps~\cite{kostavelis2015semantic} and semantic scene understanding~\cite{krizhevsky2012imagenet,sermanet2013overfeat, long2015fully, pinheiro2015learning} may provide additional information, but they would not account for dynamic objects in the environment. For field robots, particularly in sensory-deprived environments (\eg underwater), such maps are often nonexistent, and real-time scene understanding is still an open problem.
% Although semantic information about scenes is highly desirable for robotic applications, achieving scene understanding in real-time is still an open problem. However, recent developments in computer vision (\eg \cite{krizhevsky2012imagenet,sermanet2013overfeat, long2015fully, pinheiro2015learning}) provide tools to formulate potential solutions.
% It is obvious that semantic information about scenes is helpful, but it has been a challenging task for a robot to understand its environment. 
% However, recent developments in computer vision have made significant progress in addressing this problem. Since convolutional neural networks (CNNs)~\cite{krizhevsky2012imagenet} were introduced in 2012 for object classification, CNNs have been applied to identify and localize objects: object detection~\cite{sermanet2013overfeat}, semantic segmentation~\cite{long2015fully}, and instance segmentation~\cite{pinheiro2015learning}.

SOAR uses instance segmentation, which identifies specific object instances for each pixel in the visual scene, and fuses it with depth (\ie distance) information to provide semantically-aware obstacle information to obstacle avoidance modules. We adopt YOLACT~\cite{bolya2019yolact} as the instance segmentation module for our pipeline, as shown in Fig. \ref{fig:pipeline}, because it is the first state-of-the-art model to run in real-time with reasonable accuracy. While useful, object detection is not appropriate for this task as bounding boxes generated by these algorithms will contain spurious information from scene background and other objects. While semantic segmentation approaches are useful, they do not discriminate between object instances, and this information is needed for the proposed approach of semantically-aware object avoidance.

In this paper, we make the following contributions:
\begin{compactitem}
    \item Propose a pipeline for combining depth estimation and instance segmentation for semantic obstacle avoidance,
    \item Develop a semantically-aware obstacle avoidance algorithm to keep flexible distances from objects,
    \item Create an instance segmentation dataset of underwater obstacles to train an instance segmentation model, and
    \item Demonstrate the efficiency of the proposed pipeline in both underwater and terrestrial simulated environments.    
\end{compactitem}

\section{RELATED WORK}
\subsection{Instance Segmentation}
Research in object detection has studied models to improve accuracy while keeping real-time inference speed since the appearance of YOLO~\cite{Redmon_2016_CVPR}, one of the first real-time object detection models. However, instance segmentation poses more complex challenges, and achieving good accuracy in real-time has been difficult. %Therefore, research in instance segmentation has been focused on developing models with improved accuracy but without real-time inference.
% Since the introduction of real-time object detection model YOLO \cite{Redmon_2016_CVPR} in 2015, researchers have focused on developing real-time object detection models. However, unlike object detection models, the research for instance segmentation has been focused on improving accuracy.
FCIS~\cite{li2017fully} is the first end-to-end CNN-based model for instance segmentation. It is built on R-FCN~\cite{dai2016r} and utilizes position-sensitive inside/outside score maps to generate instance segmentation proposals. Mask R-CNN~\cite{he2017mask}, which is an extension of Faster R-CNN~\cite{ren2015faster}, performs segmentation in a two-stage process by generating Region of Interest (RoI) proposals first and then creating a mask based on the RoI from the first stage. PANet~\cite{liu2018path} improves the accuracy of segmentation from Mask R-CNN by enriching information propagation. MS R-CNN~\cite{huang2019mask} outperforms Mask R-CNN by adding a \textit{MaskIoU head} to align the scores of the masks. Although the aforementioned models show accurate results, their two-stage-based structures make real-time instance segmentation infeasible. In order to overcome the structural problem, YOLACT~\cite{bolya2019yolact} conducts two predictions in parallel: mask prototypes and per-instance mask coefficients. Then, the predictions are combined linearly to yield masks. This allows a single-stage structure and inference in real-time with reasonable accuracy.
% \cite{liu2018path} PAnet
% \cite{huang2019mask} MS RCNN
% \cite{he2017mask} Mask RCNN
% \cite{li2017fully} FCIS
% \cite{bolya2019yolact} Yolact
% \input{figs/pipelinenew}
% \input{tables/result0}

% is critical for safe navigation in autonomous robots. Given its importance, there is a large body of work

\subsection{Obstacle Avoidance}
Obstacle avoidance, unsurprisingly, has seen significant development (\eg \cite{rahman2019svin2,xanthidis2020navigation,manderson2020vision}) given its importance in safe robot navigation. Here, we focus specifically on sensor-based approaches where no information about the environment is available beyond what is received from sensors (see \cite{hoy2015algorithms} for a complete discussion). Sensor-based approaches typically plan a short-horizon trajectory at every time step~\cite{bircher2016receding, hoy2012collision}. A classic obstacle avoidance technique is the Artificial Potential Field, first proposed by \cite{khatib1986real}. This technique assigns artificial repulsive fields to obstacles and attractive fields to goals, thereby guiding the robot toward a goal while simultaneously avoiding obstacles. Other approaches include vector field histograms (VFH) \cite{borenstein1991vector}, receding horizon control \cite{ogren2002provably}, and Voronoi diagrams \cite{masehian2004voronoi}.

Most obstacle avoidance that incorporates semantic information is focused on developing socially-aware responses to human obstacles, \eg  \cite{foka2010probabilistic, sisbot2007human, ziebart2009planning}. Similar to our approach, \cite{sisbot2007human} instructs the robot to avoid humans more than inanimate objects. However, their work is focused on path planning in mapped environments and uses model-based methods to estimate the human's location.

Another approach for obstacle avoidance in marine robotics, based on conditional imitation learning, is presented in \cite{manderson2020vision}. This approach uses data collected from expert users to learn what navigational action to take given an input image, but does not explicitly model different behaviors for different types of obstacles.

% \subsection{Active Perception}
% \label{active_percept}
% A robot can actively perceive the environment by choosing actions that enhance or otherwise contribute to its perception and understanding. The idea of active perception was first formalized in \cite{bajcsy1988active}. A robot's actions can directly influence the information it senses (for example, adjusting camera exposure in response to light; touching an object to test a hypothesis about its identity). Therefore, actions and perceptions are tightly coupled, and active perception encompasses the study of how agents can make decisions that allow them to perceive the most relevant information from their environment. See \cite{bajcsy2018revisiting} for a more thorough discussion of the field. 
%We are particularly interested in indirect object search, where the agent adjusts its viewpoint in order to perceive the desired target. %performs an easy search for an object semantically related to a target object \cite{riazuelo2013roboearth} 

% Active Perception (Active Vision specifically) is defined as a study of Modeling and Control strategies for perception. 

% \input{figs/pipelinenew2}
\begin{figure}
    \vspace{3mm}
    \centering
    % \missingfigure[figwidth=0.49\textwidth]{test}
    \includegraphics[width=\linewidth]{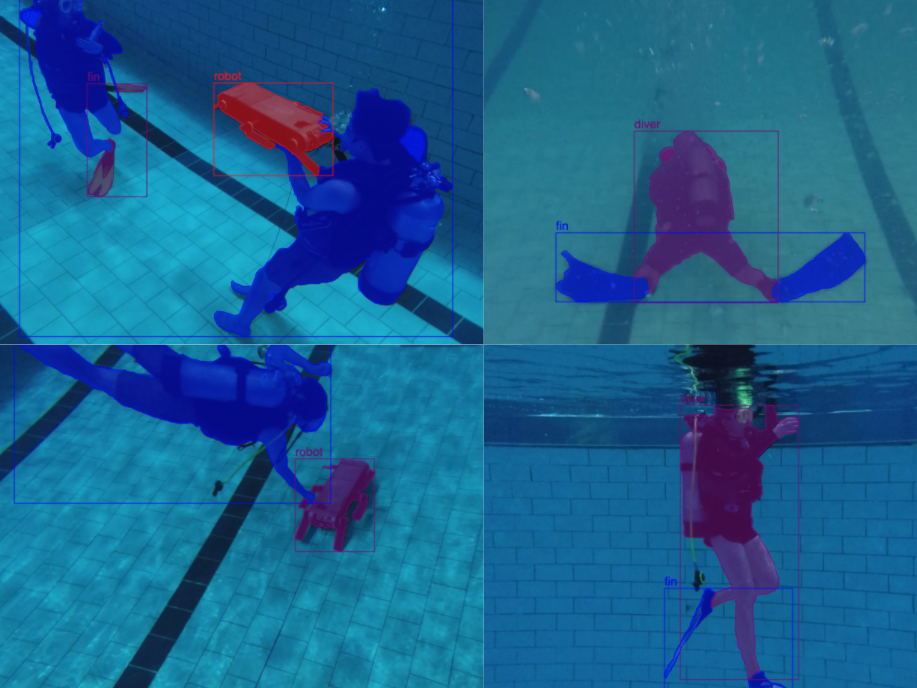}
    \caption{Examples of labeled training images from our underwater dataset showing three classes: \textit{diver, robot}, and \textit{fin} (the colors for each class are randomly selected per image).}
    \label{fig:soarlabel}
  \vspace{-4mm}
\end{figure}
\section{METHODOLOGY}
\label{sec:methodology}
The proposed approach incorporates both instance segmentation and depth information to intelligently avoid obstacles in unstructured and dynamic environments, with the goal to optimize robot paths (\eg in terms of distance traveled, energy spent, and time taken) without compromising obstacle avoidance capabilities. The system obtains object labels and pixel locations from instance segmentation and fuses the information with depth information to assign clearance distances to obstacles. % objects are obstacles that need to be avoided. 

%The system can be instructed to avoid some classes of obstacles more than others, and it can also be instructed to ignore certain classes of obstacles.

% The system can also use the instance segmentation information to make decisions about how to explore the environment. Our approach assigns different berths to each object class and makes navigation decisions with the goal of exploring areas of the environment while minimizing the travel time.

\subsection{Information Fusion using a Stereo Camera}
We use a stereo camera as our only sensor to acquire (1) pixel-level masks and labels of each object using instance segmentation, and (2) depth information. Once both are acquired, we fuse them to provide semantic information to an obstacle avoidance module (see Fig.~\ref{fig:pipeline}).

\subsubsection{Instance Segmentation with Transfer Learning}\label{sec:insseg}
We choose YOLACT~\cite{bolya2019yolact} as the base instance segmentation module due to its real-time inference and competitive accuracy. We use ResNet50-FPN as a backbone network for achieving maximum inference speed since robots are likely to use low-power computation units (\eg an Nvidia Jetson TX2) to perform semantic inference. We collect a total of $2,263$ images, of which $1,682$ are labeled with \textit{diver}, \textit{robot}, and \textit{fin} (\ie diver's flippers) classes; see Fig.~\ref{fig:soarlabel} for training images labeled using the Supervisely~\cite{supervisely} tool. In addition, we use $581$ images from the SUIM dataset~\cite{SUIM} which has \textit{diver}, \textit{fish}, and \textit{robot} classes. We refine a pre-trained YOLACT model, initially trained with the MS COCO dataset~\cite{lin2014microsoft}, with this additional data.

% 581 images for transfer learning
\subsubsection{Depth Estimation}
Our depth estimation process is as follows: %OpenCV~\cite{opencv_library} is used for  
\begin{enumerate}
\item We perform stereo rectification to obtain a transform matrix $\mathbf{R}$, projection matrix $\mathbf{P}$, and disparity-to-depth mapping matrix $\mathbf{Q}$ for each camera using a camera matrix $\mathbf{K}$ and distortion parameters $\mathbf{D}$ from each camera, a rotation matrix $\mathbf{R}$ between the first camera coordinate and the second camera coordinate, and a translation vector $\mathbf{T}$ between two cameras.
\item Next, we remove distortion from each image using the $\mathbf{K}$,  $\mathbf{D}$, $\mathbf{R}$, and $\mathbf{P}$ matrices.
\item After rectifying each pair of images, we run stereo matching to generate a disparity map.
\item Lastly, we estimate the depth information from the disparity map using $\mathbf{Q}$. 
% disparity using stereo matching (StereoSGBM\_create)
% generating depth map (reprojectImageTo3D)
\end{enumerate}

\begin{figure}
    \vspace{3mm}
    \centering
    \scalebox{0.7}{\input{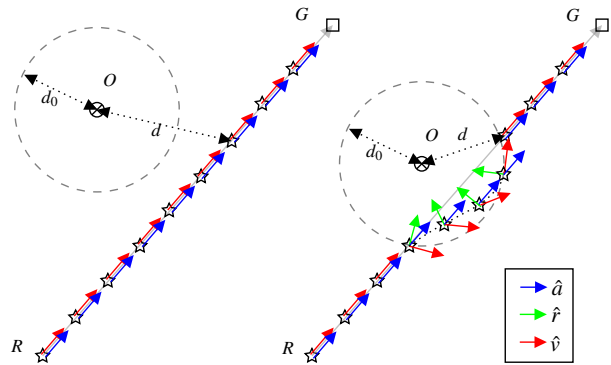}}
    \caption{Example trajectories of a robot around an obstacle using the proposed approach. The repulsive potential $\hat{r}$ affects the robot only if $d \leq d_0$. The position of object $O$ is represented by a cross, the goal position $G$ by a square, and the position of robot $R$ over time with stars. (left): when the obstacle is not along the direct path from the robot to the goal, it does not affect robot navigation. (right): the robot's direct path to the goal brings it closer than $d_0$ distance to the obstacle; our algorithm forces the robot to circumnavigate around it.}
    \label{fig:vector}
    \vspace{-4mm}
\end{figure}

\begin{figure*}[!t]
\begin{subfigure}{.5\textwidth}
  \centering
  % include third image
  \includegraphics[width=.99\linewidth]{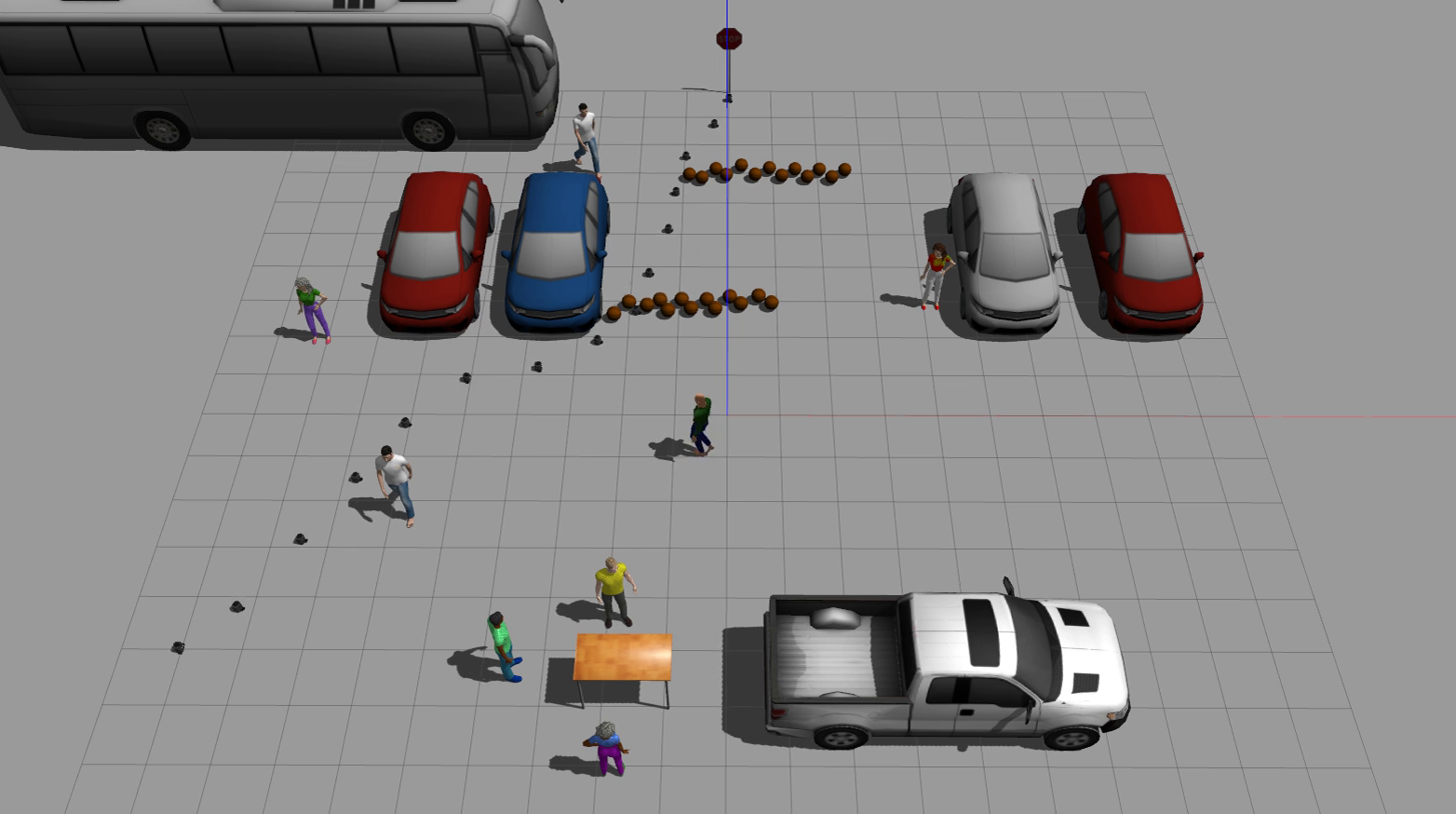}  
  \caption{Terrestrial exploration with SOAR.}
  \label{fig:turtlesense}
\end{subfigure}
\begin{subfigure}{.5\textwidth}
  \centering
  % include fourth image
  \includegraphics[width=.99\linewidth]{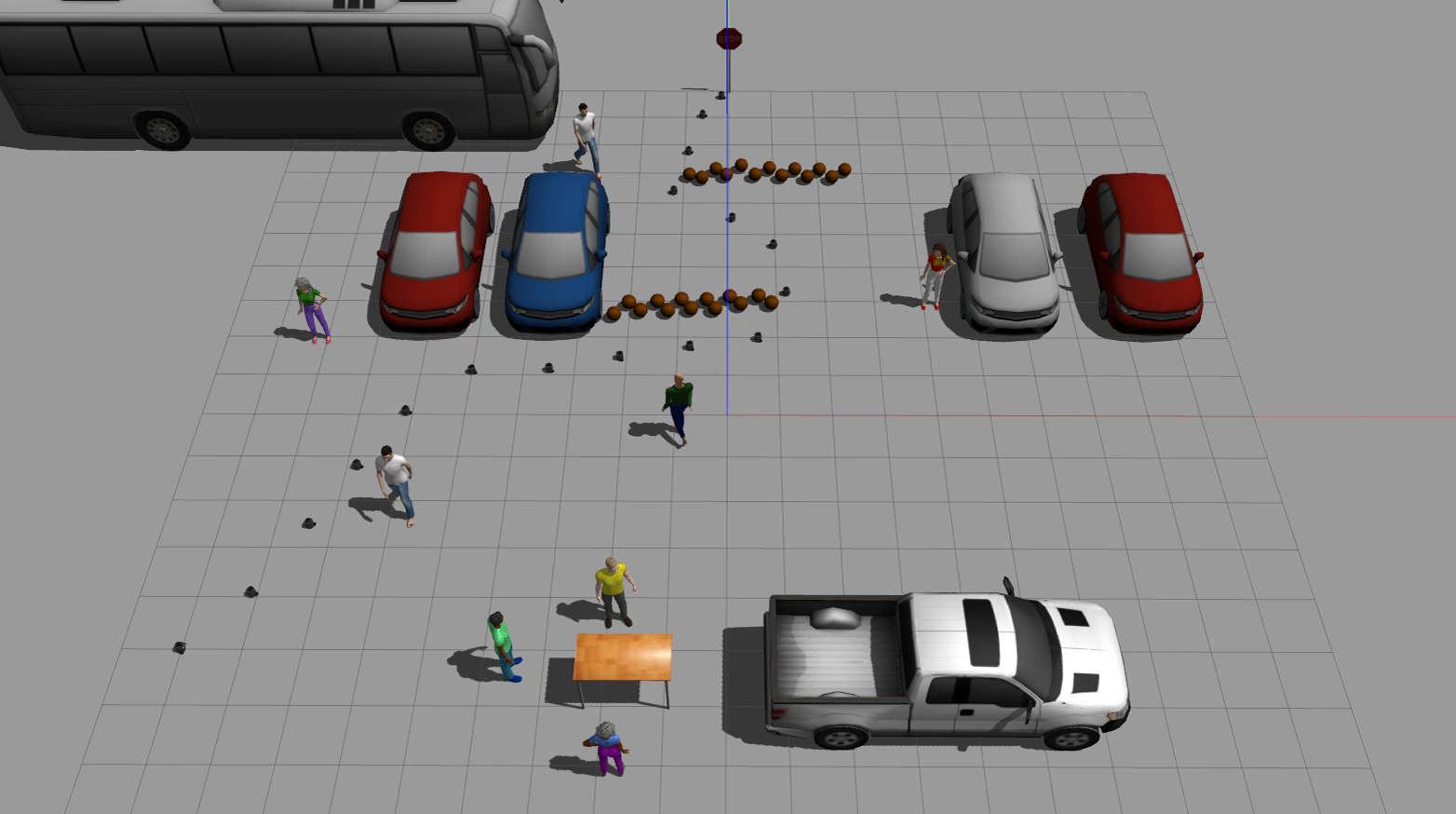}  
  \caption{Terrestrial exploration without SOAR.}
  \label{fig:turtlenosense}
\end{subfigure}

\begin{subfigure}{.5\textwidth}
  \centering
  % include first image
  \includegraphics[width=.99\linewidth]{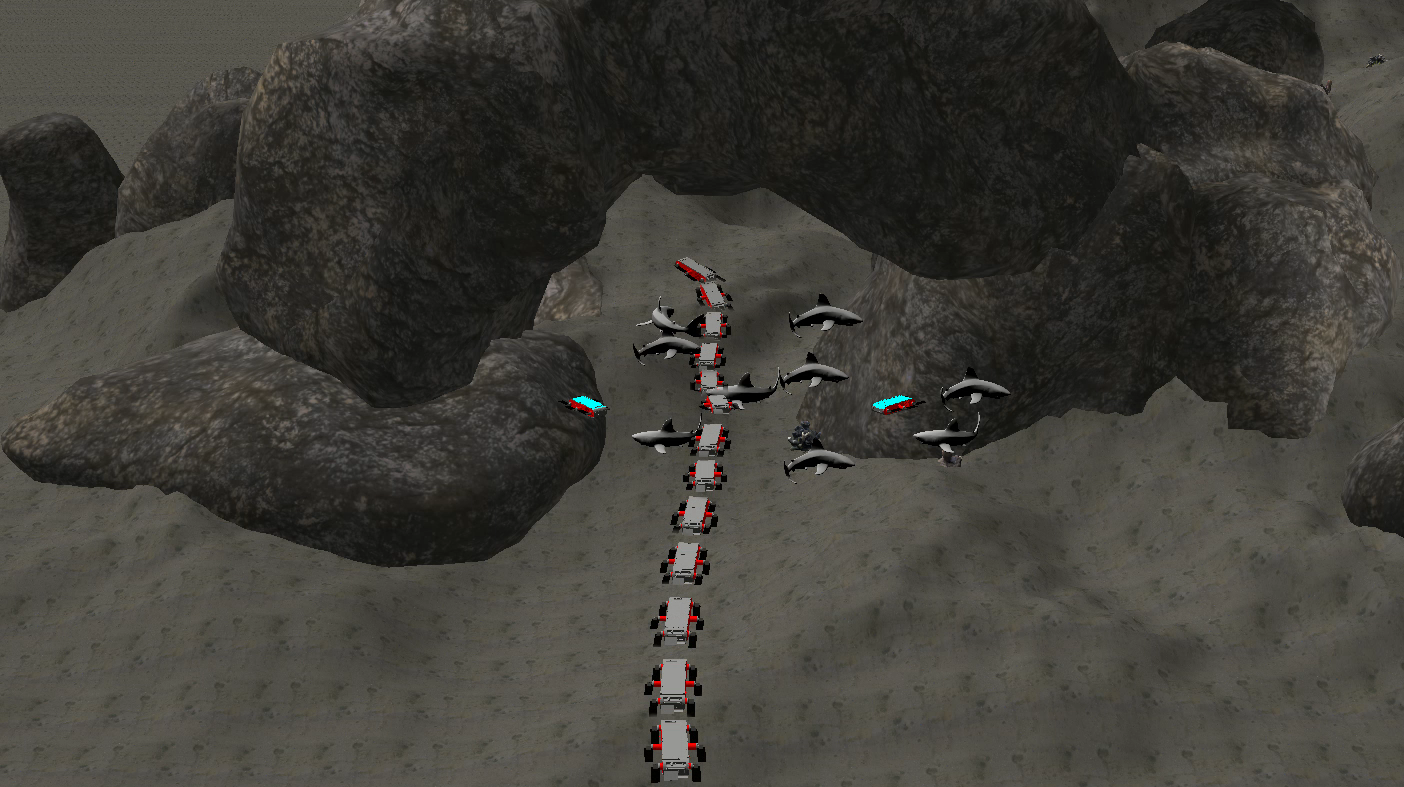}  
  \caption{Underwater exploration with SOAR.}
  \label{fig:gazeboaquasense}
\end{subfigure}
\begin{subfigure}{.5\textwidth}
  \centering
  % include second image
  \includegraphics[width=.99\linewidth]{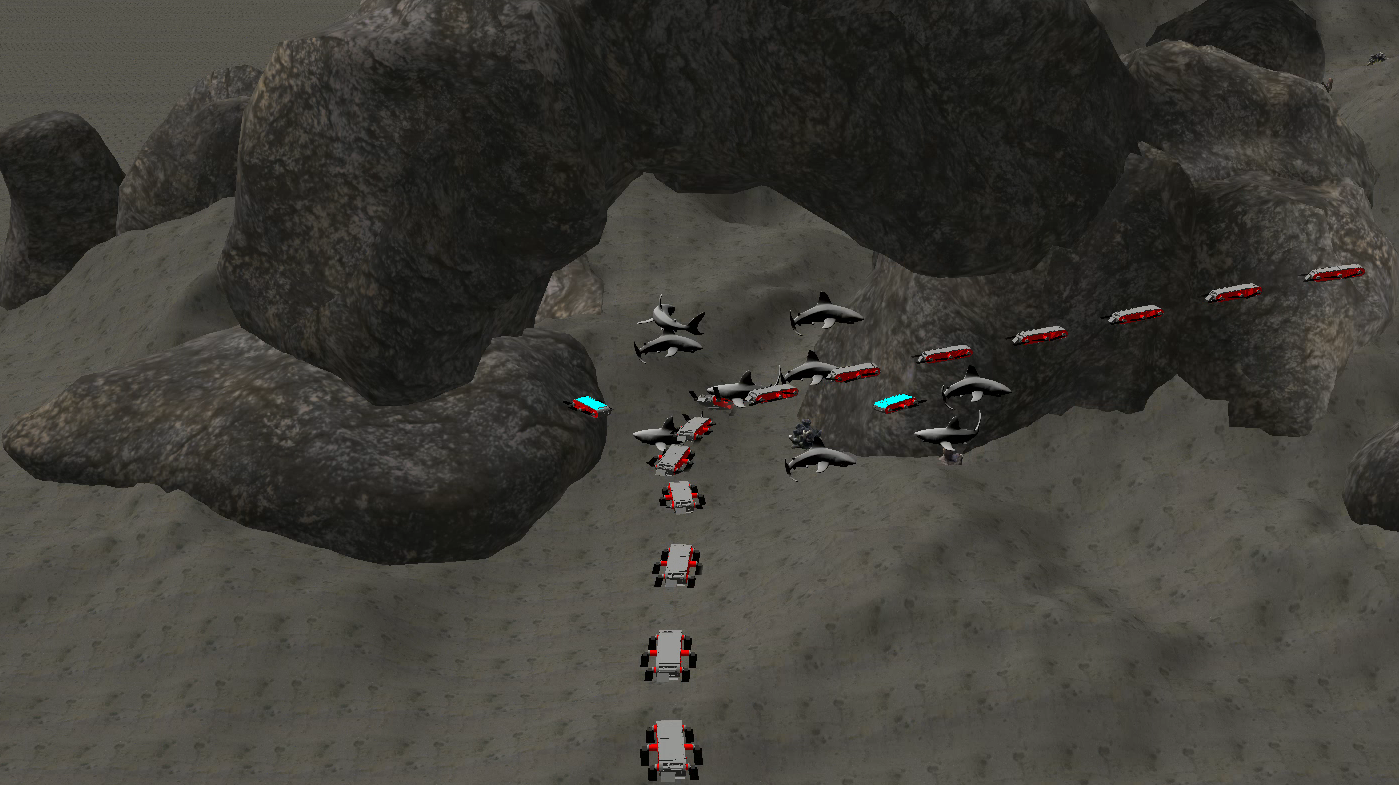}  
  \caption{Underwater exploration without SOAR.}
  \label{fig:gaezboaquanosense}
\end{subfigure}
\caption{Samples from the terrestrial and underwater simulation tests. In Fig. \ref{fig:turtlesense} and \ref{fig:turtlenosense}, the Turtlebot moves from the bottom left corner to the upper middle area. In Fig. \ref{fig:gazeboaquasense} and \ref{fig:gaezboaquanosense}, Aqua moves from the lower middle point to the middle of the arch. Fig. \ref{fig:turtlesense} and \ref{fig:gazeboaquasense} show that SOAR finds more efficient paths to explore environments while avoiding significant obstacles.}
\label{fig:gazebo}
\vspace{-4mm}
\end{figure*}

\subsection{Obstacle Avoidance} %and Active Exploration}
Our obstacle avoidance algorithm is inspired by the Artificial Potential Field (APF) method~\cite{khatib1986real}. APF uses an attractive potential $f_a$ to guide the robot toward the goal and a repulsive potential $f_r$ to push the robot away from obstacles. The attractive potential is calculated as: %shown in Eq. \ref{eq:fa} and \ref{eq:fr}. 

\begin{equation} \label{eq:fa}
    f_a(x) = c(|x - x_g|)^2
\end{equation}

Here $c$ is a scaling constant, $x$ is the current robot position, and $x_g$ is the goal position. The repulsive potential is calculated as
\begin{equation} \label{eq:fr}
f_r(x) = 
\begin{cases}
    \eta(\frac{1}{p(x)} - \frac{1}{d_0})^2 & \text{if $p(x) \leq d_0$} \\ 
                                      0 & \text{if $p(x) > d_0$}
\end{cases}
\end{equation}

where $\eta$ is a constant, $p(x)$ is the closest obstacle to the position $x$, and $d_0$ is the largest distance from the obstacle at which the robot can sense the obstacle's repulsive force. 

In our approach, we determine $d_0$ based on semantic information about the obstacles. For instance, we assign $d_0 = 0$ for objects we can ignore (\eg bubbles or sports balls) while assigning a larger value for the objects (\eg coral reef, robots, people) we intend to avoid. In our approach, unlike APF, the robot navigates around the boundary of an obstacle, keeping a constant distance of $d_0$ from the obstacle (Fig~\ref{fig:vector}). The circumnavigation behavior is similar to the bug-2 navigation algorithm \cite{lumelsky1987path}. The robot, however, may face the challenge of maintaining fixed distances (\ie $d_0$) from obstacles when circumnavigating due to errors in state estimation and external forces (\eg surge for underwater robots operating in open waters and wind for aerial robots).

We introduce two unit vectors, $\hat{a}$ and $\hat{r}$, to implement the circumnavigation with the concept of attractive and repulsive potentials from APF. $\hat{a}$ points from the robot towards the goal and $\hat{r}$ points from the robot to the obstacle. With the two vectors, we update the robot's direction of movement $\hat{v}$ at any given point as follows:

\begin{equation}\label{eq:v}
    \hat{v} = 
\begin{cases}
    \hat{a} & \text{if $|x_o - x| > d_0$} \\ 
    % \hat{a} + c_1\hat{r} & \text{if $|x_o - x| = d_0$} \\
    \hat{a} + c_1c_2\hat{r} & \text{if $|x_o - x| \leq d_0$}
\end{cases}
\end{equation}

where $c_1$ is defined as a negative dot product between $\hat{a}$ and $\hat{r}$, and $c_2$ is an additional factor to keep the distance $d_0$ between the robot and the obstacle.

\begin{equation} \label{eq:c1}
    c_1 = -\hat{a} \cdot \hat{r}    
\end{equation}

We introduce the constant $c_1$ to make $\hat{v}$ perpendicular to $\hat{r}$ when the distance between the robot and the obstacle is $d_0$. However, due to the robot's momentum, the robot may still approach closer than $d_0$ to the obstacle. We use the additional factor $c_2$ to scale the repulsive component $c_1\hat{r}$ and to enforce distance $d_0$ from the obstacle:

%drive the robot away if the robot moves within $d_0$ of the obstacle.

\begin{equation} \label{eq:c2}
    c_2 = \frac{1 - b}{d_0}|x_o-x| + b     
\end{equation}

Here $x_o$ is the obstacle position, $x$ is the robot position, and $b$ is a constant greater than 1 that represents the maximum value $c_1\hat{r}$ can be scaled by. $c_2$ scales inversely with the robot's distance to the obstacle, and obtains a value between $1$ and $b$ as we approach the obstacle.

% \input{figs/alg2_fig}
% \scalebox{0.7}{\input{figs/alg2}}

% \input{tables/results}
% \input{tables/yolactresults}
% \input{tables/soarresults}
\begin{table*}[t!]
\vspace{2mm} 
     \caption{Instance segmentation results (mAP) trained on our underwater dataset}
        \centering
        \begin{tabular}{c|c|c|c|c|c|c|c|c|c|c|c}
        \hline
        &all& .50 &.55 &.60 &.65 &.70 &.75&.80 &.85 &.90 &.95 \\
        \hline 
        box&71.14&91.96&90.29&89.48&87.89&84.09&79.54&75.20&62.56&39.07&11.29 \\
        mask&69.38&93.72&93.51&92.14&90.47&85.64&81.71&69.39&55.94&29.55&1.71\\
       \hline
       \end{tabular}
     \label{tab:yolactresults}
\end{table*}
\section{EXPERIMENTS AND RESULTS}
The ongoing COVID-19 pandemic prohibited field trial validations of the proposed method. However, we use realistic simulation using ROS Gazebo~\cite{ROS2009ICRA} worlds for both terrestrial and underwater cases to validate our algorithm. %We choose goal points for each case with one condition: a robot should be able to see the goal points if there is no obstacle between the robot and the goal points. In this way, we can test if the robot can travel to visible points (local navigation) without GPS. 
Because our goal is local navigation of unstructured, mapless environments, we choose goal points for each case with one condition: the goal point should be something visible to the robot when the robot is at its starting position if there is no obstacle between the robot and the goal points.
Additionally, we use a mobile GPU (Nvidia Jetson TX2) with a stereo camera (Intel RealSense) to evaluate the performance of our model to mimic realistic robotic hardware.%Additionally, we conduct bench tests to quantify the performance of our algorithm on an embedded GPU (Nvidia Jetson Nano).

\subsection{Simulated Terrestrial Trials}
We have created a terrestrial world in Gazebo, simulating a parking lot environment. The scene was chosen to mimic a robot attempting delivery or curbside pickup from a departmental store, a relatively common occurrence in many parts of the world under the Coronavirus pandemic. The world has various types of objects, including \textit{sports balls, cars, buses, people walking} or \textit{standing}, and \textit{tables}. We evaluate our model on a workstation equipped with an Intel i5-8600K CPU and an Nvidia GTX 1080 GPU. We add a stereo camera to a simulated Turtlebot robot to estimate depth and infer instance segmentation. To validate our algorithm, we intentionally block the shortest path from a robot to a goal point with \textit{sports balls}, as shown in Fig. \ref{fig:turtlesense} and \ref{fig:turtlenosense}, since we select the \textit{sports ball} category as an object the robot can safely run into (\ie a \textit{not-an-obstacle} object). The Turtlebot starts from the bottom left corner of the world and aims to reach the stop sign at the top. We use pre-trained COCO weights with YOLACT for instance segmentation. We use the modified APF-based obstacle avoidance algorithm as described in Section~\ref{sec:methodology} under two conditions: receiving semantic information about obstacles (SOAR) and without receiving any semantic information about obstacles (non-SOAR).
%To validate our strategy, we tested a classic APF-based obstacle avoidance module under two conditions: receiving semantic information about obstacles (test), and without receiving any semantic information about obstacles (control)
To measure the effectiveness of the semantic obstacle avoidance approach, we measure the travel time from a starting point to a goal point to evaluate each model's performance by running 10 tests for each case. We also note the path chosen by the robot in each of the SOAR and non-SOAR cases. 
% Both our model and a conventional obstacle avoidance-based exploration model have been tested 10 times, and we measured the travel time from a starting point to a goal point to evaluate each model's performance.

\subsection{Simulated Underwater Trials}
Our underwater world has been designed to mimic the ocean floor environment, including corals, rocks, and fauna. The world includes \textit{fish, robots}, and an \textit{underwater rock arch formation}. As shown in Fig. \ref{fig:gazeboaquasense} and \ref{fig:gaezboaquanosense}, the \textit{arch} is blocked by \textit{fish} and \textit{robots} to test the efficacy of the SOAR approach compared to non-SOAR. We select the \textit{fish} category as a \textit{not-an-obstacle} object class. We simulate the  Aqua AUV~\cite{dudek2007aqua}, equipped with a stereo camera, to test our model with both SOAR and non-SOAR algorithms. The robot starts from the bottom of the world and aims to travel through the arch to the other side of the rock formation. We use the instance segmentation model trained on our own dataset (as mentioned in Section~\ref{sec:insseg}) with the four classes (\ie \textit{diver, robot, fish}, and \textit{fin}). The hardware and trial configurations are unchanged from that of the terrestrial case.

% We also use a mobile GPU (NVIDIA Jetson Nano) with a stereo camera (Intel RealSense) to evaluate the performance our model to mimic realistic robotic hardware.

\subsection{Results}
We use both quantitative and qualitative evaluation to test our semantic obstacle avoidance approach, as shown in Fig. \ref{fig:gazebo} and Table \ref{tab:soarresults}. Vision algorithms are implemented using the OpenCV~\cite{opencv_library} library.
% \begin{figure*}
% \begin{subfigure}{.5\textwidth}
%   \centering
%   % include third image
%   \includegraphics[width=.99\linewidth]{imgs/turtleseg.jpg}  
%   \caption{..}
%   \label{fig:turtleseg}
% \end{subfigure}
% \begin{subfigure}{.5\textwidth}
%   \centering
%   % include fourth image
%   \includegraphics[width=.99\linewidth]{imgs/turtledisparity.jpg}  
%   \caption{..}
%   \label{fig:turtledis}
% \end{subfigure}

% \begin{subfigure}{.5\textwidth}
%   \centering
%   % include first image
%   \includegraphics[width=.99\linewidth]{imgs/aquaseg.jpg}  
%   \caption{..}
%   \label{fig:aquaseg}
% \end{subfigure}
% \begin{subfigure}{.5\textwidth}
%   \centering
%   % include second image
%   \includegraphics[width=.99\linewidth]{imgs/aquadisparity.jpg}  
%   \caption{..}
%   \label{fig:aquadis}
% \end{subfigure}
% \caption{Samples from the terrestrial and underwater simulation tests...}
% \label{fig:segdis}
% \end{figure*}

\begin{figure}
\begin{subfigure}{.49\linewidth}
  \centering
  % include third image
  \includegraphics[width=\linewidth]{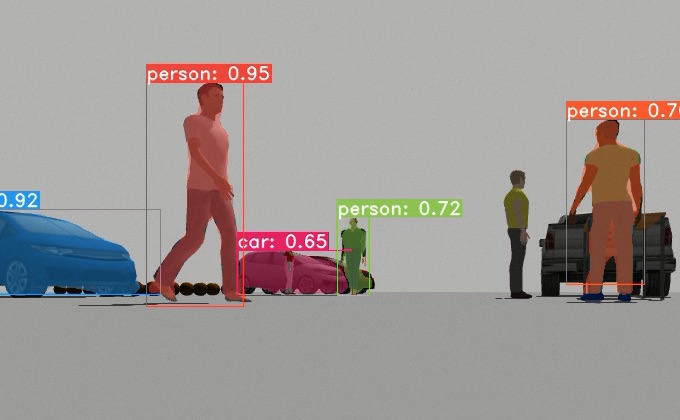}  
  \caption{Turtlebot : Segmentation}
  \label{fig:turtleseg}
\end{subfigure}%
\begin{subfigure}{.49\linewidth}
  \centering
  % include fourth image
  \includegraphics[width=\linewidth]{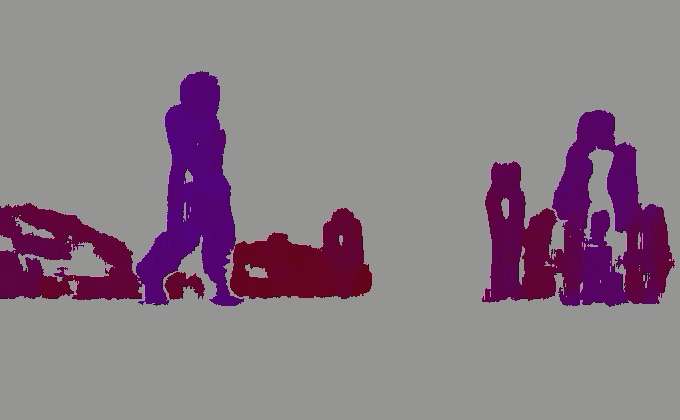}  
  \caption{Turtlebot : Disparity}
  \label{fig:turtledis}
\end{subfigure}

\begin{subfigure}{.481\linewidth}
  \centering
  % include first image
  \includegraphics[width=\linewidth]{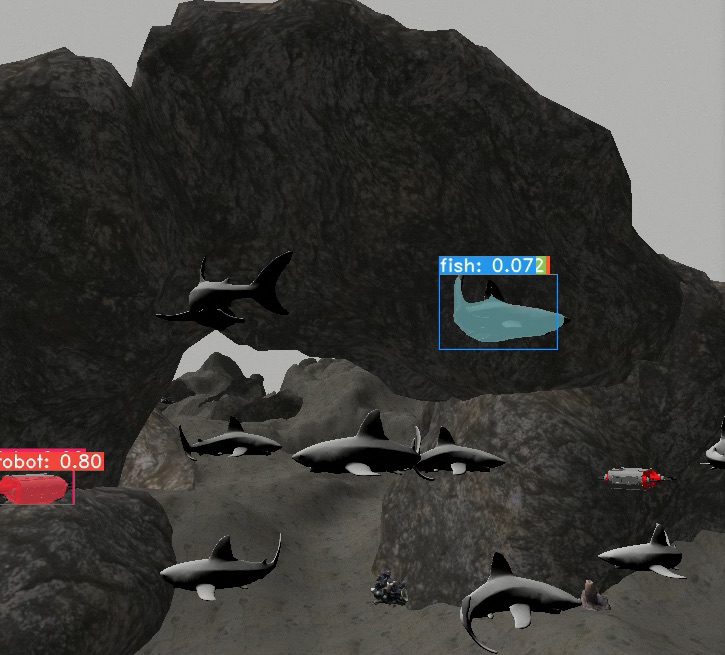}  
  \caption{Aqua : Segmentation}
  \label{fig:aquaseg}
\end{subfigure}
\begin{subfigure}{.48\linewidth}
  \centering
  % include second image
  \includegraphics[width=\linewidth]{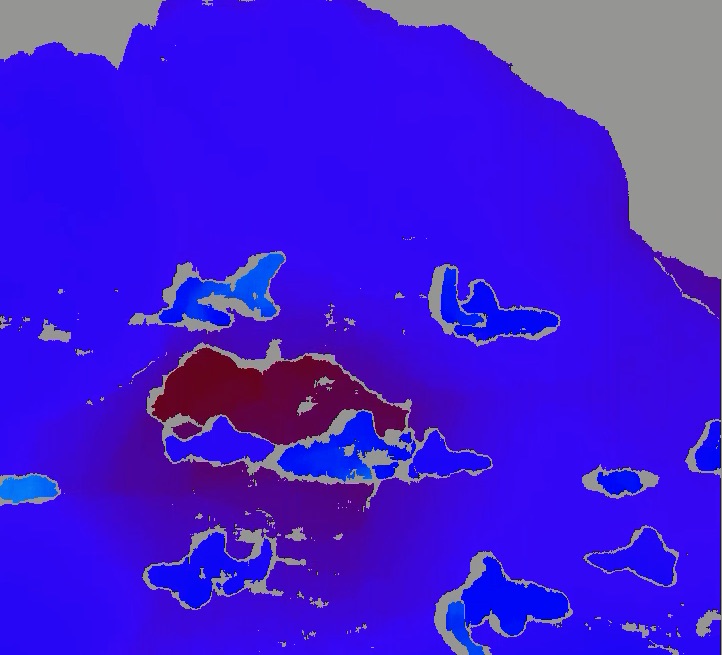}  
  \caption{Aqua : Disparity}
  \label{fig:aquadis}
\end{subfigure}
\caption{Instance segmentation and disparity estimation from the view of robots. Fig. \ref{fig:turtleseg} and \ref{fig:turtledis} are captured from the Turtlebot in the terrestrial world, and Fig. \ref{fig:aquaseg} and \ref{fig:aquadis} are from the Aqua robot in the underwater world.}
\label{fig:segdis}
% \vspace{-5mm}
\end{figure}

% \begin{figure}
% \centering
%         \begin{subfigure}[b]{0.475\textwidth}
%             \centering
%             \includegraphics[width=\textwidth]{imgs/turtleseg.jpg}
%             \caption[Network2]%
%             {{\small Network 1}}    
%             \label{fig:mean and std of net14}
%         \end{subfigure}
%         \hfill
%         \begin{subfigure}[b]{0.475\textwidth}  
%             \centering 
%             \includegraphics[width=\textwidth]{turt}
%             \caption[]%
%             {{\small Network 2}}    
%             \label{fig:mean and std of net24}
%         \end{subfigure}
%         \vskip\baselineskip
%         \begin{subfigure}[b]{0.475\textwidth}   
%             \centering 
%             \includegraphics[width=\textwidth]{Example-Image}
%             \caption[]%
%             {{\small Network 3}}    
%             \label{fig:mean and std of net34}
%         \end{subfigure}
%         \hfill
%         \begin{subfigure}[b]{0.475\textwidth}   
%             \centering 
%             \includegraphics[width=\textwidth]{Example-Image}
%             \caption[]%
%             {{\small Network 4}}    
%             \label{fig:mean and std of net44}
%         \end{subfigure}
%         \caption[ The average and standard deviation of critical parameters ]
%         {\small The average and standard deviation of critical parameters: Region R4} 
%         \label{fig:mean and std of nets}
% \end{figure}
\subsubsection{Instance Segmentation}
%We have labeled 1,682 images and use 581 images from the SUIM dataset. This leaves us 2,263 total images for training the model.
We train the instance segmentation model on a dataset of underwater imagery we collected, and also one that is openly available, as mentioned in Section~\ref{sec:insseg}.
% We collect a total of $2,263$ images, of which $1,682$ images are labeled with \textit{diver}, \textit{fish}, \textit{robot}, and \textit{fin} (\ie diver's flippers) classes. In addition, we use $581$ images from the SUIM dataset~\cite{SUIM} which has \textit{diver}, \textit{fish}, and \textit{robot} classes. 
We train for $800,000$ epochs which took five days on an Nvidia Titan XP GPU. Table \ref{tab:yolactresults} shows the average mAP (Mean Average Precision) score over IoU (Intersection over Union) thresholds from $0.50$ to $0.95$. With higher accurate localization (higher IoU thresholds), the mAP values decrease. Overall, the mAP values of bounding boxes are slightly higher than those of segmentation masks.

\subsubsection{Simulated Terrestrial Trials}
We achieve $\approx 20$ \textit{fps} while simultaneously running simulation and algorithms on the Nvidia Titan XP GPU. Table \ref{tab:turtlebot} shows the average travel time (in simulation time) from the starting point to the goal for both SOAR and non-SOAR algorithms. Although both algorithms can reach the goal, the non-SOAR algorithm takes $14\%$ longer time than when using SOAR. Samples from each case are shown in Fig. \ref{fig:turtlesense} and \ref{fig:turtlenosense}. Fig.~\ref{fig:turtleseg} and~\ref{fig:turtledis} show how Turtlebot understands scenes during its exploration. The SOAR algorithm reduces travel time by utilizing non-obstacle information (\ie \textit{sports ball}) obtained from the instance segmentation model. This demonstrates how instance segmentation information can assist in efficient exploration while safely avoiding obstacles. Information from bounding box detection and \textit{semantic} segmentation is not sufficient to provide detailed information for a robot to explore environments, particularly with \textit{intra-class occlusion.}

\subsubsection{Simulated Underwater Trials}
As our Gazebo world uses detailed hydrodynamic effects for the underwater simulation, we obtain $\approx 10$ \textit{fps} during our tests. Unlike the terrestrial case, the non-SOAR algorithm fails to reach the goal, as seen in Table \ref{tab:aqua}. This is because we terminate the non-SOAR algorithm in the following cases: 1) Aqua is heading in the wrong direction, 2) Aqua is stuck between rocks, and 3) it takes too long ($\geq 2$ minutes) to reach the goal. Sample cases from each scenario are captured in Fig. \ref{fig:gazeboaquasense} and \ref{fig:gaezboaquanosense}. The cyan-colored robots shown are ``obstacles'', and immobile, to simulate a moving Aqua robot (silver/red colors) avoiding other robots in the field. With the SOAR algorithm, Aqua is able to reach the goal (\ie under the \textit{arch}) by ignoring the group of fish. The system ignores them because of the semantic knowledge that fish do not present a collision danger. Fig.~\ref{fig:aquaseg} and ~\ref{fig:aquadis} show a snapshot of Aqua's view from the trials. The results show that for underwater domains, obstacle avoidance without understanding the scene could significantly extend the travel time at best and fail to reach the goal at worst.

\begin{table}[t!]
\vspace{2mm} 
     \caption{Terrestrial and Underwater Simulation Trial Results}
    \begin{subtable}[h]{0.49\textwidth}
        \centering
        \begin{tabular}{c | c|c| c|c}
        \hline
        & \multicolumn{2}{c|}{SOAR} & \multicolumn{2}{c}{non-SOAR}\\
        \cline{2-5}
         &  Travel time(s) & Goal & Travel time(s) & Goal\\
        \hline
        1 &  89 & \cmark & 99 & \cmark\\
        2 &  93& \cmark& 98& \cmark\\
        3 & 83& \cmark& 120& \cmark\\
        4 & 90& \cmark& 99& \cmark\\
        5 & 87& \cmark& 100& \cmark\\
        6 & 85& \cmark& 98& \cmark\\
        7 & 90& \cmark& 101& \cmark\\
        8 & 88& \cmark& 97& \cmark\\
        9 & 87& \cmark& 98& \cmark\\
        10 & 89& \cmark& 101& \cmark\\
        \hline
        Avg &  88.1& &101.1 & \\ 
       \hline
       \end{tabular}
       \caption{Turtlebot}
       \label{tab:turtlebot}
    \end{subtable}
        \hfill
    \\
    
    \begin{subtable}[h]{0.49\textwidth}
        \centering
        \begin{tabular}{c | c|c| c|c}
        \hline
        & \multicolumn{2}{c|}{SOAR} & \multicolumn{2}{c}{non-SOAR}\\
        \cline{2-5}
         &  Travel time(s) & Goal & Travel time(s) & Goal\\
        \hline
        1 & 74& \cmark & 97 & \xmark\\
        2 & 69& \cmark&  75& \xmark\\
        3 & 73& \cmark&  88& \xmark\\
        4 & 70& \cmark&  120& \xmark\\
        5 & 69& \cmark&  91& \xmark\\
        6 & 68& \cmark&  94& \xmark\\
        7 & 70& \cmark&  122&\xmark\\
        8 & 72& \cmark&  111& \xmark\\
        9 & 70& \cmark&  111& \xmark\\
       10 & 74& \cmark&  99& \xmark\\
        \hline
        Avg  &70.9 &  &100.8 & \\ 
       \hline
       \end{tabular}
       \caption{Aqua}
       \label{tab:aqua}
    \end{subtable}

     \label{tab:soarresults}
\end{table}

\subsubsection{Bench Test}
We achieved $\approx 5$ \textit{fps} inference speed while running the SOAR algorithm on the Jetson TX2. We expect to achieve faster inference speed in more capable mobile platforms (\eg AGX Xavier).

\subsubsection{Limitations}
When the inference produced by instance segmentation incorrectly classifies an obstacle as a non-obstacle, a collision can occur. Additionally, the inference time needs to be sufficiently fast to capture the objects' motion. In other words, if an object moves far faster than the inference speed, it could cause the obstacle avoidance algorithm to fail.

\section{CONCLUSIONS}
In this paper, we propose an obstacle avoidance algorithm that incorporates both instance segmentation and depth information to perceive its surroundings from only a pair of stereo image as input. We are able to use the instance segmentation labels to inform a robot about which visible obstacles in its environment should be either avoided or ignored. Finally, we present the SOAR algorithm as a viable way to explore unstructured environments with the obtained visual information. We validate our algorithm on both terrestrial and underwater simulations; quantitative results show that our algorithm can lead to efficient and intelligent robotic navigation decisions in unstructured environments, which can result in extending the duration of robot operations. 

We plan to extend this work in multiple directions. First, we
intend to integrate a sonar sensor with the camera to make SOAR robust to poor visibility conditions. Sonar readings will be used to provide additional information about obstacle locations. Visual data will be exploited to fine-tune a robot's motion when the robot approaches obstacles in close proximity. Additionally, we will improve instance segmentation by reducing the model size and inference time; training the model with increasing the size of the dataset; and optimizing it. Lastly, we plan to implement of the proposed algorithm on a robotic platform for actual in-the-field validation. 

% Future work will improve instance segmentation by reducing the model size and inference time, and implement of the proposed algorithm on a robotic platform for actual in-the-field validation. We also hope to achieve improved  performance by increasing the size of the dataset and optimizing the network. 

% address poor visibility here
% We also plan to integrate a sonar sensor to our algorithm to handle poor visibility conditions. Initial obstacle detection will be made via sonar sensors and cameras will be used to adjust the distances when obstacles appear in visible distances.

% A qualitative evaluation of our system on data from terrestrial and underwater environments indicates that incorporating instance segmentation and depth information can lead to rich, intelligent navigation decisions from robots in unstructured environments. Future work involves implementation on a robotic platform, tracking the movement of objects in the scene, and expanding the system's semantic knowledge base.
% \input{tex/6_appendix}
% \input{tex/7_ack}
\bibliographystyle{IEEEtran}
\bibliography{citation}

\end{document}